\pdfoutput=1

\documentclass[11pt]{article}

\usepackage[]{acl}

\usepackage{times}
\usepackage{latexsym}
\usepackage{graphicx}
\usepackage{amsmath}
\usepackage{url}
\usepackage{graphicx}
\usepackage{subfigure} 
\usepackage{multirow}
\usepackage{makecell}
\usepackage{colortbl}
\usepackage{amsfonts}

\usepackage{amssymb}
\usepackage{booktabs}
\usepackage{colortbl}
\usepackage{makecell}
\usepackage{verbatim}
\usepackage{latexsym}
\usepackage{graphicx,stackengine,scalerel}

\usepackage{stmaryrd}
\usepackage{helvet}
\usepackage{courier}
\usepackage{amsmath}
\usepackage{url}
\usepackage{multirow}
\usepackage{booktabs}
\usepackage{enumitem}
\usepackage{amssymb}
\usepackage{marvosym}
\usepackage{color}
\usepackage{caption}
\usepackage{titlesec}
\usepackage{floatrow}
\usepackage{caption}
\usepackage{rotating}
\usepackage{pifont}

\newfloatcommand{capbtabbox}{table}[][\FBwidth]

\usepackage[T1]{fontenc}

\usepackage[utf8]{inputenc}
\usepackage{verbatim}
\usepackage{enumitem}
\usepackage{array}

\usepackage{booktabs}

\usepackage{microtype}

\newcommand{\af}[1]{\textcolor{blue}{#1}}
\definecolor{darkgreen}{rgb}{0,0.35,0}

\title{SETI: Systematicity Evaluation of Textual Inference}

\author{Xiyan Fu \\
  Dept. of Computational Linguistics \\
  Heidelberg University \\
  \texttt{fu@cl.uni-heidelberg.de} \\\And
  Anette Frank \\
  Dept. of Computational Linguistics \\
  Heidelberg University \\
  \texttt{frank@cl.uni-heidelberg.de} \\}

\begin{document}
\maketitle
\begin{abstract}
We propose SETI (Systematicity Evaluation of Textual Inference), a novel and comprehensive benchmark designed for evaluating pre-trained language models (PLMs) for their systematicity capabilities in the domain of textual inference. Specifically, SETI offers three different NLI tasks and corresponding datasets to evaluate various types of systematicity in reasoning processes. In order to solve these tasks, models are required to perform compositional inference based on known primitive constituents. We conduct experiments of SETI on six widely used PLMs. Results show that various PLMs are able to solve \textit{unseen compositional inferences} when having encountered the knowledge of how to combine primitives, with good performance. However, they are considerably limited when this knowledge is unknown to the model (40-100 \% points decrease). Furthermore, we find that PLMs can improve drastically once exposed to crucial compositional knowledge in minimalistic shots. These findings position SETI as the first benchmark for measuring the future progress of PLMs in achieving systematicity generalization in the textual inference.
\end{abstract}

\section{Introduction}
Natural Language Inference (NLI) de\-ter\-mines whether a \textit{hypothesis} follows from a \textit{premise} \citep{dagan2013recognizing, bowman-etal-2015-large, williams-etal-2018-broad} and has been explored for decades. Existing large pre-trained language models (PLMs) have shown remarkable performance on this task \citep{devlin-etal-2019-bert, raffel2019exploring, lan2020albert}. To better assess the true capabilities of models to perform NLI, various associated tasks and benchmarks have been proposed.
These works concentrate on exploring how models make predictions, e.g. by establishing `hard' NLI datasets \citep{koreeda-manning-2021-contractnli-dataset} or asking models to `explain' their predictions through highlighting 
\citep{camburu2018snli}, or by generating plausible explanations \citep{bhagavatula2020abductive}. But little is known about how well such models are able to address \textit{compositional generalization}.

\begin{table}[t]
\centering
\resizebox{\columnwidth}{!}{
\begin{tabular}{@{}lcccccc@{}} \hline
\multirow{2}{*}{Task} & \multicolumn{4}{c}{Train} &\multicolumn{2}{c}{Test}  \\ \cmidrule(r){2-5} \cmidrule(r){6-7} 
& \multicolumn{2}{c}{Primitive Concepts} & \multicolumn{2}{c}{Compositional Concepts}  & \multicolumn{2}{c}{Compositional Concepts} \\ \hline
Task1&\begin{minipage}[b]{0.1\columnwidth}
		\centering
		\raisebox{-.2\height}{\includegraphics[width=\linewidth]{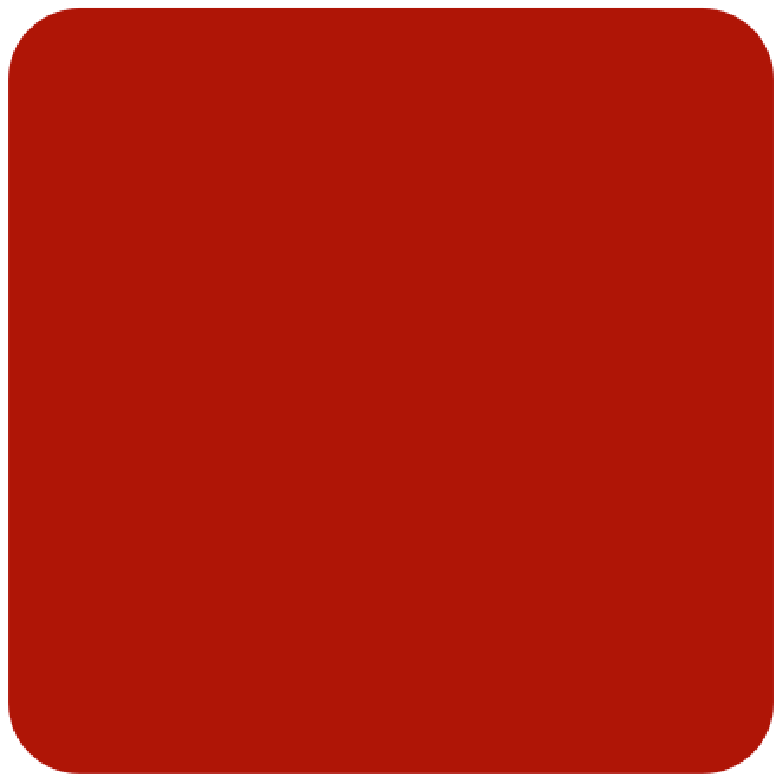}}
	\end{minipage}  & \begin{minipage}[b]{0.1\columnwidth}
		\centering
		\raisebox{-.2\height}{\includegraphics[width=\linewidth]{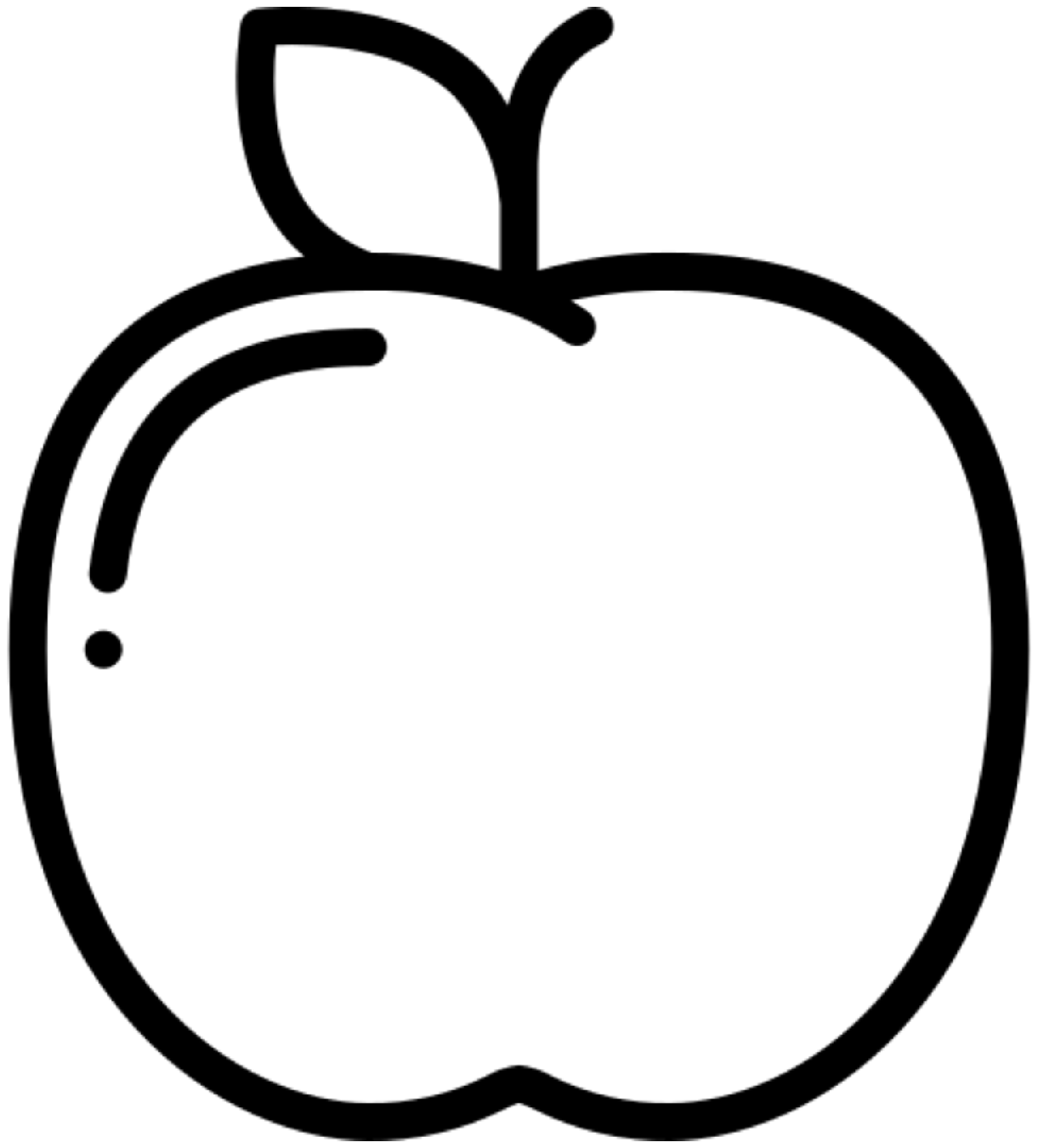}}
	\end{minipage}& & & \multicolumn{2}{c}{\begin{minipage}[b]{0.1\columnwidth}
		\centering
		\raisebox{-.2\height}{\includegraphics[width=\linewidth]{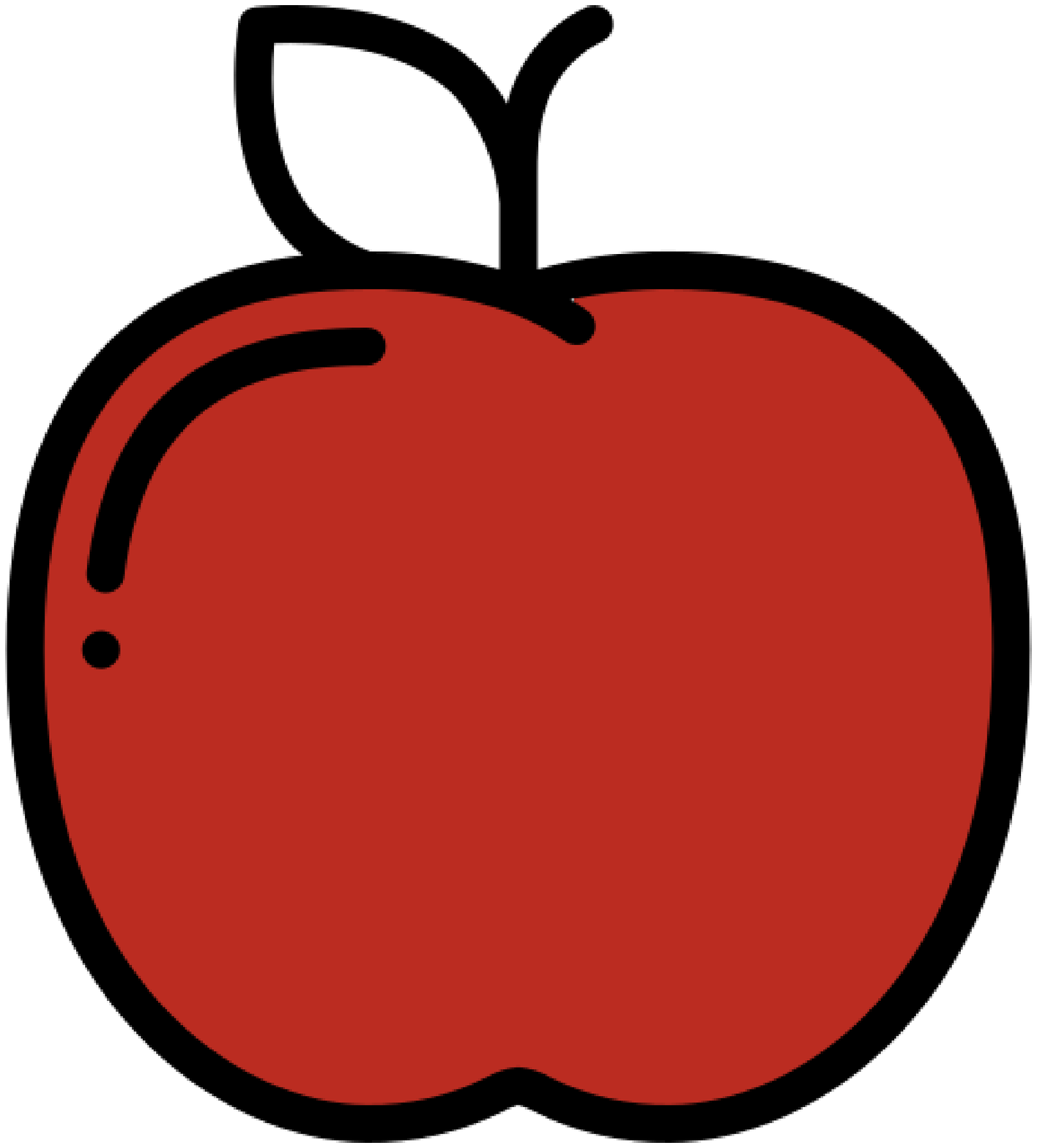}}
	\end{minipage}} \\ \hline
Task2& & &\begin{minipage}[b]{0.1\columnwidth}
		\centering
		\raisebox{-.2\height}{\includegraphics[width=\linewidth]{red_apple.eps}}
	\end{minipage} & \begin{minipage}[b]{0.1\columnwidth}
		\centering
		\raisebox{-.2\height}{\includegraphics[width=\linewidth]{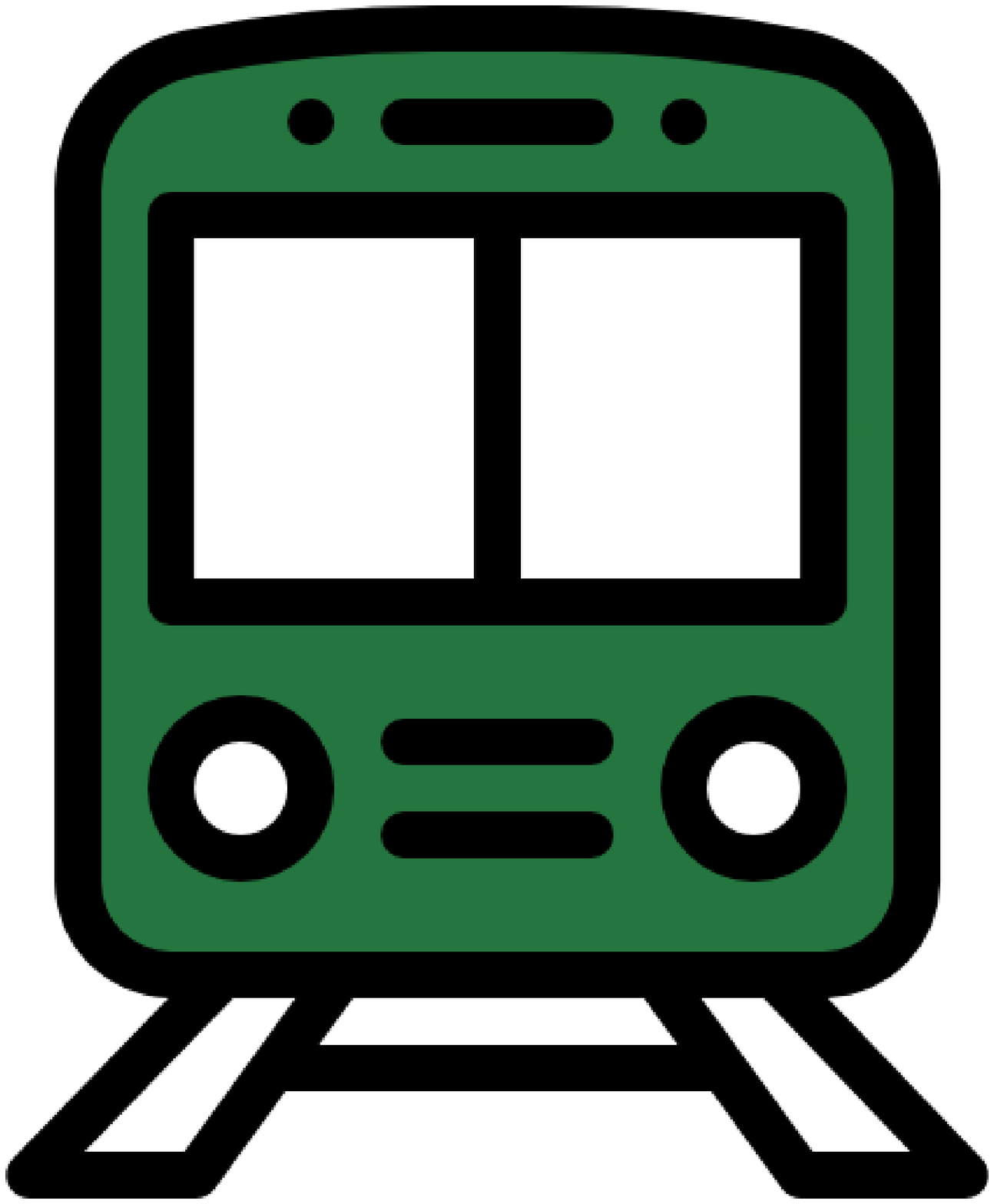}}
	\end{minipage} & \begin{minipage}[b]{0.1\columnwidth}
		\centering
		\raisebox{-.2\height}{\includegraphics[width=\linewidth]{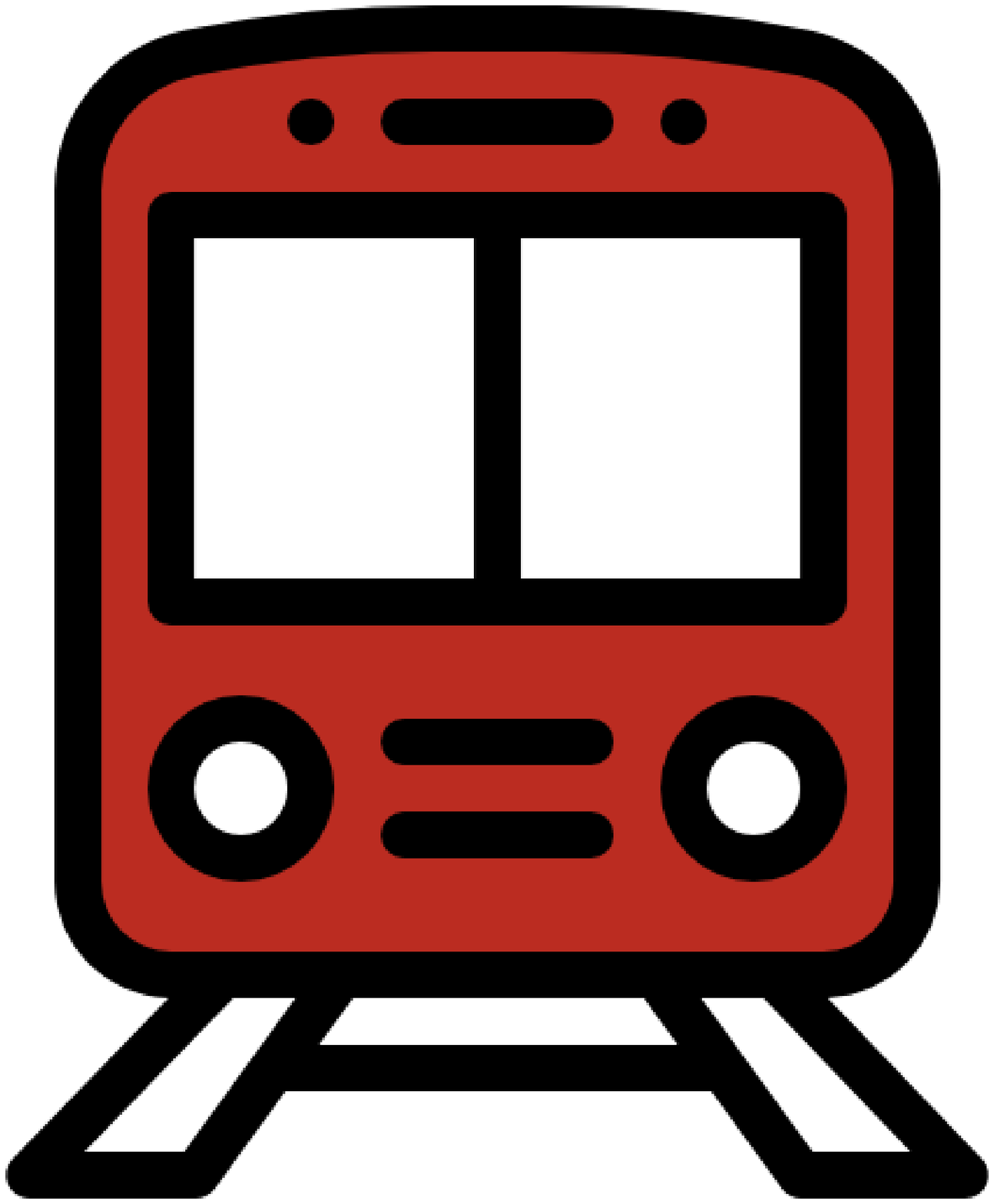}}
	\end{minipage} & \begin{minipage}[b]{0.1\columnwidth}
		\centering
		\raisebox{-.2\height}{\includegraphics[width=\linewidth]{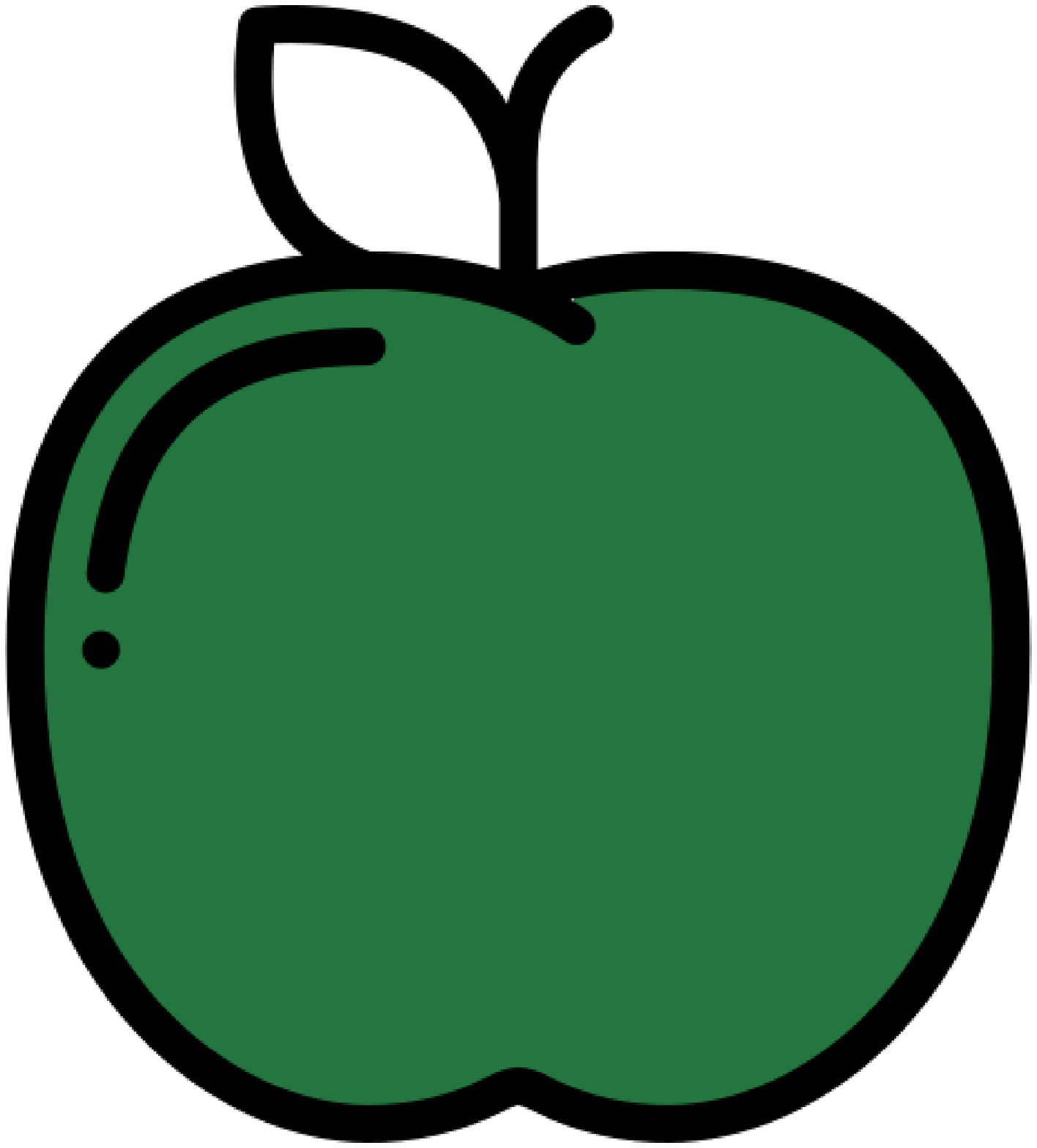}}
	\end{minipage} \\ \hline
Task3& \multicolumn{2}{c}{\begin{minipage}[b]{0.1\columnwidth}
		\centering
		\raisebox{-.2\height}{\includegraphics[width=\linewidth]{red.eps}}
	\end{minipage}} & \multicolumn{2}{c}{\begin{minipage}[b]{0.1\columnwidth}
		\centering
		\raisebox{-.2\height}{\includegraphics[width=\linewidth]{green_train.eps}}
	\end{minipage}} & \multicolumn{2}{c}{\begin{minipage}[b]{0.1\columnwidth}
		\centering
		\raisebox{-.2\height}{\includegraphics[width=\linewidth]{red_train.eps}}
	\end{minipage}} \\ \hline
\end{tabular}}
\caption{Illustrating three visual tasks realizing different forms of \textit{systematicity} in compositional generalization.}.
\label{tasks_visual}
\end{table}

Compositional generalization focuses on how to combine 
primitive units to predict larger com\-pou\-nds \citep{hupkes2020compositionality}. A key property\- un\-der\-ly\-ing compositional generalization is \textit{systematicity} \citep{fodor1988connectionism}, a hallmark of human cognition. Systematicity concerns 
the ability of (re)combining known constituents and composing rules.
For example, humans who understand `red apple' and `green train' are 
able to conceptualize `red train' by recombining `red' and `train' into
a new concept. Similar effects of systematicity (generalization) can be studied in
Natural Language Understanding (NLU) \citep{lake2018generalization, kim-linzen-2020-cogs}. Since PLMs have achieved re\-sults on par with human performance
by fitting NLI training data \citep{wang2019superglue}, we aim to evaluate to what extent
these 
models can master different types of \textit{systematicity} in textual inference.
We propose a novel benchmark SETI (\textit{Systematicity Evaluation of Textural Inference}),
which extensively explores systematicity in NLI. SETI contains three interrelated yet independent tasks covering various types of systematicity:
1) \textbf{Task1: primitives $\rightarrow$ compositions} aims to evaluate if models can perform compositional inference if primitive constituents of the given inference task
have been learned independently. 2) \textbf{Task2: compositions $\rightarrow$ compositions} aims to evaluate if models can perform novel compositional inferences if their constituents have been learned in other compositions. 3) \textbf{Task3: primitives and compositions $\rightarrow$ compositions} aims to evaluate if models can perform novel compositional inferences if one primitive constituent has been learned independently, while the other has only been encountered in compositions. SETI can be used to explore systematicity in NLI \textit{comprehensively} since it considers all possibilities of how to construct a novel composition from known constituent types, derived from the ‘permutation and combination’\footnote{While permutations make sense in a setting that deals with grammaticality, this does not hold for inference, hence we do not consider permutation order for SETI.} theory acting between primitives and compositions. We introduce these tasks in detail in Section \S \ref{model_sec}. To make the instantiations of systematicity covered in SETI easily accessible, we indicate three analogous visual tasks in Table \ref{tasks_visual}. They test if models can understand: i) a novel compositional concept \textit{red apple} -- given the primitive concepts \footnote{In the visual domain, primitive concepts such as colors and object properties seldomly occur independently of objects, instead they occur in composition with objects. Here, they are only used for task clarifications.}\textit{red} and \textit{apple} have been learned independently; ii) a novel compositional concept \textit{red train} -- given constituent concepts \textit{red} and \textit{train} have been encountered in compositions \textit{red apple} and \textit{green train}; iii) a novel compositional concept \textit{red train} -- given \textit{red} has been learned independently, and \textit{train} in the compositional concept \textit{green train}.

To apply SETI in practice, we define veridical inference \citep{karttunen1979conventional, ross-pavlick-2019-well} and natural inference as \textit{primitives}, and their combinations as \textit{compositions}.
For each systematicity task setting, we provide two instantiations: \textit{trivial} and \textit{non-trivial}, depending on the variety of instances presented to the model in
training. While both settings fulfill the given task requirements, the 
 \textit{non-trivial} setting
is more challenging because 
the compositional inference knowledge of how to combine constituents is not seen in training.


We evaluate six well-known PLMs on all SETI tasks. They
show good performance
in \textit{trivial} settings, but inferior results in \textit{non-trivial} settings, for all tasks. This indicates that models can generalize well to unseen compositions when 
constituents and compositional knowledge are known, while 
they are limited when they lack knowledge about how to compose constituents.
Hence, we further explore whether, and to what extent we can enhance the systematicity capabilities. Our experiments indicate that all PLMs benefit greatly from being exposed to minimal
doses of relevant compositional instances.


Our main contributions are as follows:

\begin{enumerate}[label=\roman*), noitemsep]
    \item We introduce SETI (\textbf{S}ystematicity \underline{E}valuation of \underline{T}extual \underline{I}nference), which to our knowledge is
    the first benchmark to comprehensively evaluate
     the systematicity capabilities of PLMs when performing NLI.
    \item We provide datasets for three NLI challenge tasks that evaluate systematicity, with controlled splits for seen vs.\ unseen information.
    \item 
    We conduct experiments for six widely used PLMs. The results indicate that models generalize well to unseen compositions if they have previously acquired relevant 
    compositional inference knowledge, but are limit\-ed when lacking such knowledge.
\end{enumerate}

\section{Related Work}

\paragraph{Textual Inference}
Natural Language Inference (NLI) involves reasoning across a \textit{premise} and \textit{hypothesis}, determining the inferential relationship holding between them
\citep{dagan2013recognizing, bowman-etal-2015-large, williams-etal-2018-broad}. As one of the major
tasks for establishing Natural Language Understanding (NLU), NLI has been widely explored for decades. Recently, large pre-trained language models \citep{devlin-etal-2019-bert, raffel2019exploring, lan2020albert} exhibit remarkable performance on NLI tasks, on par with
humans.
To better explore the true NLI capabilities of models,
various associated tasks and benchmarks have been proposed.
Some work has probed NLI models by constructing hypothesis-only baselines \citep{glockner-etal-2018-breaking, liu-etal-2020-hyponli}, finding that models capture undesired biases. 
\citet{mccoy-etal-2019-right, zhou-bansal-2020-towards, gubelmann-etal-2022-philosophically} reveal that models rely on heuristics, e.g., lexical overlap, subsequence heuristics, etc. \citet{nie-etal-2020-adversarial, chien2020adversarial} evaluate models in adversarial settings and show robustness improvement by training on additional
adversarial data. Others focus on explainable NLI, such as highlighting input words that are essential for the label\citep{camburu2018snli},
or generating plausible explanations \citep{bhagavatula2020abductive}. In this work, we focus on exploring the compositional generalization abilities of PLMs when performing textual inference.

\paragraph{Systematicity} 
Systematicity is a crucial property of compositionality, which was first introduced in cognitive science \citep{fodor1988connectionism} and recently formalized in \citet{hupkes2020compositionality}. 
It is the ability to make use of known concepts to pro\-duce novel concept combinations that have not been encountered before. Recently, systematicity has been widely explored in domains such as image caption generation \citep{nikolaus-etal-2019-compositional}, visual attribute recognition \citep{misra2017red, li2020symmetry}, question answering \citep{keysers2020measuring, liu-etal-2022-challenges} and semantic parsing \citep{lake2018generalization, finegan-dollak-etal-2018-improving, kim-linzen-2020-cogs, zheng-lapata-2022-disentangled}. In this work, we focus on systematicity in the domain of textual inference.


Existing works that evaluate systematicity in textual inference only focus on one specific type. For example, \citet{yanaka-etal-2021-exploring} evaluates systematicity by testing the transitivity of inference relations.
Others conduct experiments on novel compositions involving specific linguistic phenomena, such as systematicity of predicate replacements and embedding quantifiers \citep{yanaka-etal-2020-neural}, systematicity 
when combining
lexical entailment and negation \citep{geiger-etal-2020-neural}, and systematicity of quantifiers, negation and concerning the order between premises and hypotheses \citep{goodwin-etal-2020-probing}. 

Compared to prior work, we propose a \textit{comprehensive} systematicity evaluation benchmark SETI, which: 
i) covers the full spectrum of systematicity; ii) evaluates various PLMs; and ii) showcases
how PLMs can overcome limitations in systematicity. 

\section{Reasoning Tasks for Systematicity}
\label{model_sec}
{We now define primitive and compositional in\-fe\-ren\-ces and 
introduce three NLI systematicity tasks.

\subsection{Primitive and Compositional Inferences}
\begin{table*}[ht]
\centering
\resizebox{\columnwidth}{!}{
\begin{tabular}{@{}llll@{}} \hline
\multicolumn{3}{c}{Primitive Inference Types} & Examples (premise $\rightarrow$ hypothesis)  \\ \hline
\multirow{2}{*}{\makecell[l]{Veridical\\ Inference $PI_{ver}$}} & \multicolumn{2}{l}{veridical $f_{v}(s) \rightarrow s$} & Someone realizes that a man is eating a pizza $\rightarrow$ A man is eating a pizza\\
& \multicolumn{2}{l}{non-veridical $f_{nv}(s) \nrightarrow s$} & Someone hopes that a man is eating a pizza $\nrightarrow$ A man is eating a pizza \\ \hline
\multirow{4}{*}{\makecell[l]{Natural \\ Inference $PI_{nat}$}}& \multirow{2}{*}{\makecell[l]{lexically-based \\ inference rule $\xrightarrow{lex}$}} &entailment $s \xrightarrow{lex} s'$ & A boy is jumping into the water $\rightarrow$ A kid is jumping into the water \\  
& &non-entailment $s \longarrownot\xrightarrow{lex} s'$ & A woman is smiling $\nrightarrow$ A man is smiling \\ \cmidrule(r){2-4}
& \multirow{2}{*}{\makecell[l]{structure-based \\ inference rule $\xrightarrow{stru}$}} &entailment $s \xrightarrow{stru} s'$ & The detective follows a man $\rightarrow$ A man is being followed by the detective \\
& &non-entailment $s \longarrownot\xrightarrow{stru} s'$ & A fish is being sliced by a man $\nrightarrow$ A cat is jumping into a box \\ \hline
\end{tabular}}
\caption{Examples of \textit{primitive} veridical ($PI_{ver}$) and natural ($PI_{nat}$) inferences. $s$, $s'$ represent distinct sentences. }
\label{primitive_tasks}
\end{table*}

\begin{table*}[ht]
\centering
\resizebox{\columnwidth}{!}{
\begin{tabular}{llllp{7.5cm}} \hline
$PI_{ver}$& $PI_{nat}$ &$CI_{ver\_nat}$ ($CI$) & Composed Rules & Examples (premise $\rightarrow$ hypothesis) \\ \hline
$f_{v}(s) \rightarrow s$ & $s \xrightarrow{lex} s'$ &$f_{v}(s) \xrightarrow{lex^{+}} s'$ & \ding{172} True $\wedge$
 True $\rightarrow$  True  & He realizes a boy is jumping into the water $\rightarrow$ A kid is jumping into the water    \\
$f_{v}(s) \rightarrow s$ &$s \longarrownot\xrightarrow{lex} s'$ &$f_{v}(s) \longarrownot\xrightarrow{lex^{-}} s'$ & \ding{173} True $\wedge$ False $\rightarrow$ False &  He realizes a woman smiling $\nrightarrow$ A man is smiling\\
$f_{nv}(s) \nrightarrow s$ &$s \xrightarrow{lex} s'$ &$f_{nv}(s) \longarrownot\xrightarrow{lex^{+}} s'$& \ding{174} False $\wedge$ True $\rightarrow$  False &He hopes a boy is jumping into the water $\nrightarrow$ A kid is jumping into the water   \\
$f_{nv}(s) \nrightarrow s$ &$s \longarrownot\xrightarrow{lex} s'$ &$f_{nv}(s) \longarrownot\xrightarrow{lex^{-}} s'$ & \ding{175} False $\wedge$ False $\rightarrow$  False &He hopes a woman is smiling $\nrightarrow$ A man is smiling \\ \hline
\end{tabular}}
\caption{Examples of \textit{compositional} inferences $CI$ obtained by combining veridical and natural inference (we use $\xrightarrow{lex}$ as example; $\xrightarrow{stru}$ works analogously). For $CI$, the label ($\rightarrow$/$\nrightarrow$) is decided by the Boolean `Composed Rules'. We use $lex^{+}$ and $lex^{-}$ to indicate the label of its $PI_{nat}$ component being \textit{True} or \textit{False}, respectively.}
\label{composition_tasks}
\end{table*}

Among various textual inference types, we select \textit{veridical inference} and \textit{natural inference} as two primitive inference tasks\footnote{We adopt binary labels (entailment/non-entailment) by grouping contradiction and  neutral as non-entailment.}, since they can be flexibly scaled to compositional inferences. Table \ref{primitive_tasks} shows relevant notation and corresponding examples of the two primitive inference types. 
 
\textbf{Veridical Inference} Veridical inference is strongly determined
by the lexical meaning
of sentence embedding verbs. In the context of a \textit{veridical} verb we can infer that the proposition it takes as complement is taken to hold true. By contrast, in the context of a \textit{non-veridical} verb, we \emph{can not} infer that the proposition it takes as complement is taken to hold true.
\citep{karttunen1971implicative, ross-pavlick-2019-well}. 
$PI_{ver}$ in Table \ref{primitive_tasks} shows examples of both verb classes. The verb ``realize'' in the premise ``Someone realizes that a man is eating pizza'' is \textit{veridical} in relation to the embedded proposition ``A man is eating pizza'', since speakers cannot say the premise unless they believe the latter proposition to be true. In contrast, ``hope'' is \textit{non-veridical}, since the premise ``Someone hopes that a man is eating pizza'' does not license the equivalent conclusion towards  
the hypothesis ``A man is eating pizza''.
In our work, we emphasize veridicality in verb-complement constructions and formulate their inference potential in an NLI setting, as premise-hypothesis pairs, as established by \citet{ross-pavlick-2019-well}. Specifically, the premises of all veridical inference samples follow the template ``Someone $f_{v}/f_{nv}$ that $s$'', where $f_{v}$ and $f_{nv}$ represent veridical and non-veridical complement embedding verbs, respectively. We denote samples of entailed vs.\ non-entailed veridical inferences as $f_{v}(s) \rightarrow s$ and $f_{nv}(s) \nrightarrow s$, respectively.

\textbf{Natural Inference} A pair of sentences is considered a true entailment if we are able to infer the hypothesis based on the premise. $PI_{nat}$ in Table \ref{primitive_tasks} shows examples. We categorize natural inference samples into two groups: 1) \textit{lexically-based inferences} typically build on lexical inference knowledge captured in lexical meaning relations, e.g., hypernymy \textit{boy $\rightarrow$ kid} in ``A boy is jumping into the water'' $\rightarrow$ ``A kid is jumping into the water''. 2) \textit{structure-based inferences} involve structural changes, e.g., from active to passive voice and vice versa, as in ``The detective follows the man'' $\rightarrow$ ``The man is being followed by the detective''. We restrict natural inferences to these two types to facilitate controlled data creation.
We denote entailed and non-entailed samples from these two groups as:  $s \xrightarrow{lex} s'$, $s \longarrownot\xrightarrow{lex} s'$ and $s \xrightarrow{stru} s'$, $s \longarrownot\xrightarrow{stru} s'$. 

\textbf{Composing veridical and natural inference}
To evaluate the compositional generalization ability of models, we construct \textit{compositional inferences} $CI_{ver\_nat}$ (CI) by combining \textit{primitive} veridical inference $PI_{ver}$ and natural inference $PI_{nat}$, following \citet{yanaka-etal-2021-exploring} (see Table \ref{composition_tasks}).

For such compositions to be valid, the hypothesis of a veridical inference must match the premise of a natural inference. This matching condition serves as a crucial link to perform transitive inference. Table \ref{composition_tasks} shows how a compositional inference sample \textit{`He realizes a boy is jumping into the water'} $\rightarrow$ \textit{`A kid is jumping into the water'} is composed from $PI_{ver}$ \textit{`He realizes a boy is jumping into the water'} $\rightarrow$ \textit{`\underline{A boy is jumping into the water}}' and $PI_{nat}$ \textit{`\underline{A boy is jumping into the water}}' $\rightarrow$ \textit{`A kid is jumping into the water'}. This reasoning process we denote 
as: $f_v(s) \rightarrow s$ $\wedge$ $s \rightarrow s'$ $\Rightarrow$ $f_v(s) \rightarrow s'$.

In this way, we construct four types of compositional inferences $CI$ 
from primitive $PI_{ver}$ and $PI_{nat}$ inferences, where Boolean logical rules (Table \ref{composition_tasks}, col. 3) decide the label of $CI$, i.e., whether it yields entailment or non-entailment. In case both veridical 
$PI_{ver}$ and natural inference $PI_{nat}$ resolve to \textit{True}, $CI$ yields entailment, given the Boolean logic rule \textit{True $\wedge$ True $\rightarrow$ True} (rule \ding{172}). By contrast, if $PI_{nat}$ yields non-entailment, the compositional veridical inference $CI$ will fail (rule \ding{173}). However,
compositional inference with non-verdical verbs invariably yields non-entailment,
no matter whether $PI_{nat}$ resolves to \textit{True} or \textit{False}. This is again due to Boolean logic (rules \ding{174}, \ding{175}): \textit{False $\wedge$ (True $\vee$ False) $\rightarrow$ False.} In conclusion, the first two cases of CI are more complex, since models need to follow
Boolean logic,
while a model could exploit shortcuts and invariantly predict non-entailment with non-entailing verbs in $PI_{ver}$.

\begin{table*}[ht]
\centering
\resizebox{0.8\columnwidth}{!}{
\begin{tabular}{@{}lll@{}} \toprule
\multicolumn{2}{c}{Tasks} & Examples (premise $\rightarrow$ hypothesis)  \\ \midrule
\multirow{4}*{\rotatebox{90}{Task1}}& \multirow{2}*{$D_{train}$} &$PI_{x}:$ Someone realizes a boy is jumping into the water $\rightarrow$ A boy is jumping into the water\\ 
&&$PI_{y}:$ A boy is jumping into the water $\rightarrow$ A kid is jumping into the water \\ \cmidrule(r){2-3} 
&$D_{test}$& $CI: $ Someone realizes a boy is jumping into the water $\rightarrow$ A kid is jumping into the water \\ \midrule
\multirow{4}*{\rotatebox{90}{Task2}}& \multirow{2}*{$D_{train}$} &$CI_{x}:$ Someone realizes a boy is jumping into the water $\rightarrow$ A kid is jumping into the water\\ 
&& $CI_{y}:$ Someone hopes a woman is eating a pizza $\nrightarrow$ A man is eating a pizza \\ \cmidrule(r){2-3} 
&$D_{test}$& $CI: $ Someone hopes a boy is jumping into the water $\nrightarrow$ A kid is jumping into the water \\ \midrule
\multirow{4}*{\rotatebox{90}{Task3}}& \multirow{2}*{$D_{train}$} &$PI:$ A man is driving a car $\rightarrow$ A car is being driven by a man\\ 
&& $CI:$ Someone realizes a boy is jumping into the water $\rightarrow$ A kid is jumping into the water \\ \cmidrule(r){2-3} 
&$D_{test}$& $CI: $ Someone realizes a man is driving a car $\rightarrow$ A car is being driven by a man \\\bottomrule
\end{tabular}}
\caption{Examples of three systematicity tasks from SETI. For each task, we select one sample from the \textit{trivial} setting for representation. }
\label{task_exm}
\end{table*}

\subsection{SETI Tasks}
Having characterized the two types of primitive inferences we will use in our experiments, along with ways of composing them, we will now spell out i) how to define increasingly difficult generalization tasks targeting systematicity, with ii) appropriate specifications of train and test settings, to guarantee proper assessment of a model's generalizing capacities. Table \ref{task_exm} presents examples. \par

\textbf{Task1: primitives $\rightarrow$ compositions} aims to evaluate whether a model can perform a compositional inference $CI$ if its (primitive) constituent inferences $PI_{x}$ and $PI_{y}$, have been learned independently, while their combination is unseen in training. Hence, \textit{Train and Test sets} (D$_{train|test}$) consist of instances $e$ and $\tilde{e}$:
\begin{equation}
\begin{aligned}
   D_{train} &= \{e \mid  e \in PI_{x} \vee e \in PI_{y}\}\\
   D_{test} &= \{\tilde{e} \mid \tilde{e} \in CI\}
\end{aligned}
\end{equation}
We select \textit{veridical} inference and \textit{lexically-based} natural inference as primitive inferences, and 
combinations of these two primitives as compositional inferences, as formally specified below:
\begin{equation} \label{ex:T2_nontrivial}
\begin{small}
\begin{aligned}
   PI_{x} &= PI_{ver} = \{f_{v}(s) \rightarrow s, f_{nv}(s) \nrightarrow s\} \\
   PI_{y} &= PI_{lex} = \{s \xrightarrow{lex} s', s \longarrownot\xrightarrow{lex} s'\} \\
   CI &= \{f_{v}(s) \xrightarrow{lex^{+}} s', f_{v}(s) \longarrownot\xrightarrow{lex^{-}} s', \\
   &\quad f_{nv}(s) \longarrownot\xrightarrow{lex^{+}} s', f_{nv}(s) \longarrownot\xrightarrow{lex^{-}} s'\}
\end{aligned}
\end{small}
\end{equation}
Here, sentences ($s$ and $s'$) of composed inferences $CI$ are constrained to match sentences of their primitive constituents $PI_{ver \vee lex}$. This is a \textbf{trivial} setting since the challenge is restricted to classifying compositional inference from seen primitive inferences.

However, overlaps of words between $PI_{nat}$ and $CI$ bear a risk of shortcuts \cite{sanchez-etal-2018-behavior}. Hence, we also evaluate compositional inferences in a \textbf{non-trivial} setting, where sentences used in compositional
inferences 
in $D_{test}$ are constrained to differ from sentences used in 
primitive constituents in $D_{train}$. This 
is 
doable if we guarantee 
that instances from $PI_{nat}$ and $CI$ share the same inference rules $lex_{x}$. For example, we provide `A \textit{boy} is jumping into the water $\rightarrow$ A \textit{kid} is jumping into the water' in $PI_{nat}$; and `Someone $f_v/f_{nv}$ a \textit{boy} is playing in the mud $\rightarrow$ A \textit{kid} is playing in the mud' in $CI$. In this way, models can retain the knowledge of $PI_{lex}$ by using the same inference rules, e.g., rule $x:$ \textit{boy $\rightarrow$ kid}, while we inhibit
shortcuts by using different contexts in the test set.


\textbf{Task2: compositions $\rightarrow$ compositions} aims to evaluate if a model is able to predict unseen compositional inferences $CI_{test}$ whose constituting primitives
have been encountered in other compositional inferences $CI_{train}$ in training. \textit{Train and Test sets} (D$_{train|test}$) consist of instances $e$ and $\tilde{e}$:
\begin{equation}
\begin{aligned}
   D_{train} &= \{e \mid  e \in CI_{train}\}\\
   D_{test} &= \{\tilde{e} \mid \tilde{e} \in CI_{test}\}
\end{aligned}
\end{equation}
We construct specific types of compositional training instances by combining \textit{veridical} inference with \textit{lexical} natural inference, and \textit{non-veridical} with \textit{structural} natural inference, see (\ref{ex:T2_trivial}). To evaluate 
if models can generalize to novel compositions, we switch the constituents (primitive inference types) seen in training to unseen compositional inferences in testing. I.e., we evaluate \textit{veridical} inference with \textit{structural} natural inference, and \textit{non-veridical} inference with \textit{lexical} natural inference. $CI_{train}$ and $CI_{test}$ are specified
as:
\begin{equation}
\begin{small}
\label{ex:T2_trivial}
\begin{aligned}  
   CI_{train} = &\{f_{v}(s) \xrightarrow{lex^{+}} s', f_{v}(s) \longarrownot\xrightarrow{lex^{-}} s', \\
   &f_{nv}(s) \longarrownot\xrightarrow{stru^{+}} s', f_{nv}(s) \longarrownot\xrightarrow{stru^{-}} s'\} \\
   CI_{test} = &\{f_{v}(s) \xrightarrow{stru^{+}} s', f_{v}(s) \longarrownot\xrightarrow{stru^{-}} s',  \\
   &f_{nv}(s) \longarrownot\xrightarrow{lex^{+}} s', f_{nv}(s) \longarrownot\xrightarrow{lex^{-}} s'\}
\end{aligned}
\end{small}
\end{equation}
This is a \textbf{trivial} setting, given that four composition rules 
(\ding{172}\ding{173}\ding{174}\ding{175} in Table \ref{composition_tasks}) 
have been instantiated
in the training samples. The challenge is restricted to correctly classifying novel compositions from known primitives. 

To further explore if models can generalize to novel compositions based on unseen composition rules we propose a \textbf{non-trivial} setting. Here, a model must 
combine \textit{entailed veridical} inference with \textit{entailed natural} inference, and \textit{non-\-ve\-ri\-di\-cal} inference with \textit{non-entailed natural} inference. With this, only 
rules \ding{172} and \ding{175} are instantiated by the
training samples. 
In testing we confront the model with composition instances unseen in training, by switching constituents, so that we test for the unseen rules \ding{173} and \ding{174}: we compose
\textit{entailed veridical} with \textit{non-entailed natural} inference, and \textit{non-veridical} with \textit{entailed natural} inference. 
$CI_{train}$ and $CI_{test}$ are defined as: 
\begin{equation} \label{ex:T3_nontrivial}
\begin{small}
\begin{aligned}
   CI_{train} = &\{f_{v}(s) \xrightarrow{nat^{+}} s', f_{nv}(s) \longarrownot\xrightarrow{nat^{-}} s'\} \\
   CI_{test} = &\{f_{v}(s) \longarrownot\xrightarrow{nat^{-}} s', f_{nv}(s) \longarrownot\xrightarrow{nat^{+}} s'\} 
\end{aligned}
\end{small}
\end{equation}
We expected this to be an intractable challenge, since models are now required to classify novel compositions, where identical primitives have been encountered in training compositions, but the required composition rules of tested compositions are not instantiated in the training data.

\textbf{Task3: Primitives and Compositions $\rightarrow$ Compositions} aims to evaluate whether a model is able to predict an unseen compositional inference $CI_{test}$ whose one primitive inference $PI$ has been learned independently, while the other has only been encountered in a compositional inference $CI_{train}$ in training. 
Hence, \textit{Train and Test sets} (D$_{train|test}$) consist of instances $e$ and $\tilde{e}$:
\begin{equation}
\begin{aligned}
   D_{train} &= \{e \mid  e \in PI \vee e \in CI_{train}\}\\
   D_{test} &= \{\tilde{e} \mid \tilde{e} \in CI_{test}\}
\end{aligned}
\end{equation}
We could choose either veridical or natural inference as a primitive inference $PI$.
Here, we select \textit{natural inference} as the $PI$ (veridical inference works analogously). Specifically, we construct $CI_{train}$ by combining \textit{entailed veridical} inference with \textit{lexically-based} natural inference, and define \textit{structure-based} natural inference as $PI$ . To evaluate if models can generalize to novel compositional inference,
we substitute the lexically-based natural inference component in $CI_{train}$ with structure-based natural inference to form $CI_{test}$ instances, as stated below: 
\begin{equation}
\begin{small}
\begin{aligned}
   PI &=  PI_{stru} = \{s \xrightarrow{stru} s', s \longarrownot\xrightarrow{stru} s'\} \\
   CI_{train} &= \{f_{v}(s) \xrightarrow{lex^{+}} s',
   f_{v}(s) \longarrownot\xrightarrow{lex^{-}} s'\} \\
   CI_{test} &=  \{f_{v}(s) \xrightarrow{stru^{+}} s',
   f_{v}(s) \longarrownot\xrightarrow{stru^{-}} s'\}
\end{aligned}
\end{small}
\end{equation}
This is again a \textbf{trivial} setting, given that the composition rules (\ding{172}\ding{173}) required in testing have been exemplified by training samples. That is, the challenge is restricted to correctly classifying novel compositions, where their primitives and the composition rules are known.

Analogous to Task2, we introduce a further \textbf{non-trivial} setting to evaluate if models can generalize to novel compositions that test for unseen composition rules. The primitive inference $PI$ could be either veridical or natural inference. 

Hence in one variant, we choose i) \textit{veridical} inference (-$R_{ver}$) as the $PI$, and construct the training compositions by combining \textit{entailed veridical} with \textit{entailed lexical} natural inference, while the primitive inference is \textit{non-veridical} inference. For testing, we replace the veridical inference in training compositions with independent non-veridical inferences. This setting is defined below:
\begin{equation}
\begin{small}
\begin{aligned}
   PI &= \{f_{nv}(s) \nrightarrow s'\} \\
   CI_{train} &= \{f_{v}(s) \xrightarrow{lex^{+}} s'\} \\
   CI_{test} &=  \{f_{nv}(s) \longarrownot\xrightarrow{lex^{+}} s'\}
\end{aligned}
\end{small}
\label{ex:task3nontrivial}
\end{equation}
This setting should be challenging, since models are required to evaluate novel compositions that correspond to the compositional rule \ding{174}, while they have only encountered rule \ding{172} in training. 

As alternative variant ii), we choose \textit{natural inference} (-$R_{nat}$) as the primitive inference $PI$. We construct training data for compositions by combining \textit{entailed veridical} with \textit{entailed lexical} natural inference, and define \textit{non-entailed lexical} natural inference as the primitive inference. For testing, we replace the entailed lexical inference in training compositions with independent non-entailed lexical inferences. This is defined below:
\begin{equation}
\begin{small}
\begin{aligned}
   PI &= \{s \longarrownot\xrightarrow{lex} s'\} \\
   CI_{train} &= \{f_{v}(s) \xrightarrow{lex^{+}} s'\} \\
   CI_{test} &=  \{f_{v}(s) \longarrownot\xrightarrow{lex^{-}} s'\}
\end{aligned}
\end{small}
\label{ex:task3nontrivial2}
\end{equation}
This setting is challenging, since models are required to evaluate novel compositions according to rule \ding{173}, while having only seen rule \ding{172} in training.

\begin{table}[t]
\centering
\resizebox{\columnwidth}{!}{
\begin{tabular}{@{}lllcccccc@{}} \hline
\multicolumn{2}{c}{\multirow{2}{*}{Tasks}} &\multicolumn{4}{c}{Composition Generalization} & \multicolumn{3}{c}{In-distribution} \\ \cmidrule(r){3-6} \cmidrule(r){7-9}
&&type&Train  &Dev  &Test &Train  &Dev  &Test \\ \hline
\multirow{4}*{\rotatebox{90}{Task1}} &trivial &\makecell[l]{$PI_{ver}$\\$PI_{nat}$}  &\makecell[c]{1680\\1686}  &\makecell[c]{420\\422}  &63240  &3366  &842  &63240 \\ \cmidrule(r){2-9} 
 &$\neg$ trivial &\makecell[l]{$PI_{ver}$\\$PI_{nat}$}  &\makecell[c]{600\\603}  &\makecell[c]{150\\151}  &22620  &1203  &301  &22620 \\ \hline
\specialrule{0em}{1pt}{1pt}
\multirow{2}*{\rotatebox{90}{Task2}}&trivial &$CI$ &25296  &6324  &31620  &25296  &6324  &31620 \\
& $\neg$ trivial &$CI$ &25296  &6324  &31620  &25296  &6324  &31620 \\
\specialrule{0em}{1pt}{1pt}\hline
\multirow{6}*{\rotatebox{90}{Task3}}& trivial &\makecell[l]{$PI_{nat}$\\$CI$}  &\makecell[c]{480\\480}   &\makecell[c]{120\\120}   &9000  &960  &240  &9000\\ \cmidrule(r){2-9}
& \makecell[l]{$\neg$ trivial\\-$R_{ver}$} &\makecell[l]{$PI_{ver}$\\$CI$}   &\makecell[c]{9048\\9048}   &\makecell[c]{2262\\2262}  &11310  &18096  &4524  &11310 \\ \cmidrule(r){2-9}
& \makecell[l]{$\neg$ trivial\\-$R_{nat}$} & \makecell[l]{$PI_{nat}$\\$CI$}   &\makecell[c]{600\\603}   &\makecell[c]{150\\151}  &11310  &1203  &301  &11310 \\ \hline
\end{tabular}}
\caption{Statistics of \textit{compositional generalization controlled} data. $PI$ and $CI$ indicate primitive and compositional inferences, respectively. ``type'' marks the inference types used in train and dev sets.
}
\label{data_statis}
\end{table}

\begin{figure*}
    \centering
    \includegraphics[width=6.1in]{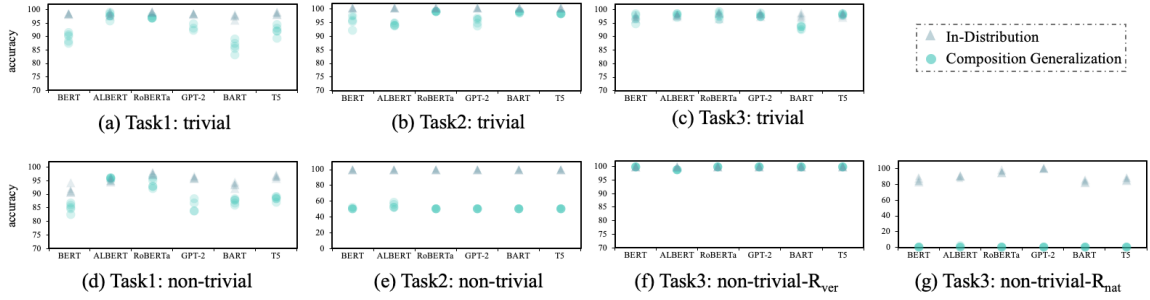}
    \caption{Performance of six PLMs on the SETI benchmark in two configurations: ``Compositional Generalization'' and ``In-Distribution''. For each task setting and PLM, we perform five runs and represent each result by a symbol.}
    \label{fig_rst_tasks}
\end{figure*}

\section{Experimental Setup}
\subsection{Dataset}
\label{data_cons}
To evaluate the systematicity capabilities 
of PLMs on the series of SETI tasks we established above, we construct controlled datasets with instances chosen from established NLI datasets. For primitive inference: 1) \textit{veridical inference}, we select 30 verbs (15 veridical, 15 non-veridical) that appear in both the MegaVeridicality2 \cite{white-etal-2018-lexicosyntactic} and the verb veridicality dataset of \citet{ross-pavlick-2019-well}, as \citet{yanaka-etal-2021-exploring} do (cf.\ Appendix.\ref{app:verbs} for details). 2) \textit{natural inference}, we extract instances from the SICK dataset \citep{marelli-etal-2014-sick} that use lexical inferences $s \xrightarrow{lex} s'$ where sentence pairs are formed from lexical relations, e.g., synonymy and hyponymy. In addition, we select structural inferences $s \xrightarrow{stru} s'$ where sentence pairs are derived from each other using the active-passive diathesis. Examples are shown in Table \ref{primitive_tasks}.
For compositional inferences, we construct instances following \S 2.1. We combine premises $f_{v/nv}(s)$ from veridical inferences with hypotheses $s'$ from natural inference. Boolean logic rules are used to assign labels for these compositional inference instances.


Based on the constructed pool of inference data, we design three \textbf{Compositional Generalization} task datasets to evaluate the systematicity of PLMs. Specifically, primitive and compositional inferences data is divided for training $D_{train}$ and testing $D_{test}$ in a controlled way, as outlined in Section \S \ref{model_sec}. This ensures that the evaluated models will be exposed to specific types of inference instances in training, while being evaluated on unseen compositional inferences. That is to say, \textit{the testing data is out of distribution from the training data}. In addition, we provide corresponding \textbf{In-Distribution} task datasets for comparison. Here, the data is divided into $D_{train}^{'}$ and $D_{test}^{'}$ by producing random splits from $D = D_{train} \cup D_{test}$. Hence, the evaluated models will, during training, encounter instances of the kind that will be presented in testing. In other words, the \textit{testing data is \textit{In-distribution} of the training data}. \textit{In-Distribution} data makes it possible to confirm whether the failure of \textit{Compositional Generalization} is due to intractable compositional inference tests or a lack of systematicity. Table \ref{data_statis} shows detailed data statistics for both configurations. For further details see
Appendix \ref{app:statistics}.

\subsection{Evaluated Models}
We choose six well-known PLMs for evaluation, of which three are masked language models (\textit{encoder-only}): BERT \citep{devlin-etal-2019-bert}, RoBERTa \citep{liu2019roberta} and ALBERT \citep{lan2020albert}; two are denoising autoencoder models (\textit{encoder-decoder}): T5 \citep{2020t5} and BART \citep{lewis-etal-2020-bart}; and one auto-regressive model (\textit{decoder-only)}: GPT-2 \citep{radford2019language}. We use standard accuracy as the evaluation metric.

For all PLMs we have chosen \textit{Large} models, with checkpoints from the Hugging Face implementation \citep{wolf-etal-2020-transformers}\footnote{\url{https://huggingface.co/models}}. We finetuned these models using the Adam Optimizer with batch size of 16. The maximum input token number is limited to 128. For each of the seven task settings, we perform five runs for each PLM, using different seeds. Further details are provided in Appendix \ref{app:models}.

\section{Experiments}
\begin{figure*}
    \centering
    \includegraphics[width=5.3in]{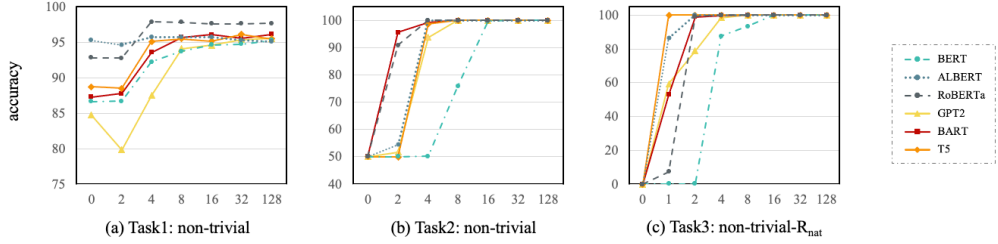}
    \caption{Few-shot performance of six well-known PLMs on three challenging sub-task of the SETI benchmark.}
    \label{fig_few-shot}
\end{figure*}

\subsection{Overview Results}
Fig.\ref{fig_rst_tasks} illustrates the performance of six well-known PLMs on the SETI benchmark across two data configurations: \textit{Compositional Generalization} and \textit{In-Distribution}. Among seven different task settings, we find the test accuracy of PLMs in \textit{In-Distribution} to be close to 100\% in most cases, with a drop to $\ge$ 80\% in Task1 and Task2, non-trivial, but always stable across five rounds. This indicates that compositional inferences of various types are feasible for the evaluated PLMs if they have seen relevant instances in training. \textit{Compositional Generalization} shows comparable results in trivial settings of Task2 (Fig.\ref{fig_rst_tasks}.b) and Task3 (Fig.\ref{fig_rst_tasks}.c), but inferior results for most of the remaining settings. This suggests that the evaluated PLMs are lacking systematicity capabilities when encountering unseen compositional inference problems, while achieving remarkable performance in \textit{In-Distribution} by fitting training data.

Comparing \textit{trivial} and \textit{non-trivial} settings across three tasks, we find: 1) In \textbf{Task1}, the test accuracy of \textit{In-Distribution} slightly decreases in the non-trivial setting, which confronts the models with novel contexts for $PI_{nat}$ inferences within the compositional test cases $CI$. This shows that the \textit{non-trivial} setting is more challenging. We also find that the performance of \textit{Compositional Generalization} drops in the non-trivial setting.
However, the encoder-only models ALBERT and RoBERTa outperform others substantially, showing strong systematicity generalization ability in both settings. 2) In \textbf{Task2}, the test accuracy of all generalization-tested models declines sharply to 50\% in the non-trivial setting, no matter how well a model performs in the trivial setting. 
And this finding holds across different rounds of each PLM, indicating that novel compositional inferences are equally challenging for all evaluated models. 
3) In \textbf{Task3}, generalization-tested models also show inferior results in the non-trivial setting, while non-trivial-$R_{ver}$ (Fig.\ref{fig_rst_tasks}.f) is an exception (100\% accuracy). This is expected
since in this setting the PLMs can solve unseen
compositional problems by exploiting
superficial characteristics during training, rather than by generalization, i.e., the capability of systematicity. Specifically, the non-trivial-$R_{ver}$ task evaluates $f_{nv}(s) \longarrownot\xrightarrow{lex^{+}} s'$ given $f_{v}(s) \xrightarrow{lex^{+}} s'$ and $f_{nv}(s) \nrightarrow s'$. In this task setting, non-veridical verbs $f_{nv}$ are only seen in non-veridical inference, which may lure models to predict non-entailment for compositional inferences containing non-veridical verbs, yet without considering the entailment class of the embedded lexical inference.

Across the different tasks, the evaluated PLMs show diverse performance for generalization testing. \textbf{Task1} is almost solved by ALBERT and RoBERTa, highlighting that some models are capable of combining different primitive inferences (learned independently) in unseen compositional inferences.
However, along with all other PLMs, none of the two remaining Tasks can be reliably solved in the controlled, non-trivial ``Compositional Generalization'' setting: i) predicting unseen compositions, the components of which have been learned during training (\textbf{Task2}) and ii) determining a novel composition, where one primitive is learned independently, while the other has been encountered in a composition during training (\textbf{Task3}).

\subsection{Few-shot Evaluation}
We conclude from the results shown in Fig \ref{fig_rst_tasks} that the evaluated PLMs are incapable of performing compositional generalization if they have not encountered crucial compositional inference knowledge during training.
Hence, we aim to explore whether, and to what extent we could 
enhance the systematicity capabilities of the evaluated PLMs,
by exposing them to small doses of relevant instances. Specifically, we select three \textit{non-trivial} sub-tasks that expect models to solve compositional inferences without encountering the required inferential knowledge in training. 
For each such task, we construct a few-shot dataset $D_{few}$ where each sample (compositional inference, CI) is constructed following \S \ref{data_cons}. 
$D_{few}$ and $D_{test}$ contain different data, i.e., $D_{test} \cap D_{few} = \emptyset$.
For each task, we evaluate few shot samples from 0 to 128, and each model is fine-tuned for three epochs. By doing so, we expect the models to learn the underlying compositional inference knowledge from the samples given in $D_{few}$, so they can finally solve $D_{test}$.

Figure \ref{fig_few-shot} shows the few-shot experiment results. Across different tasks, we find that all evaluated PLMs benefit from few-shot samples that teach the model relevant compositional inference knowledge.
In \textbf{Task1}, most PLMs show a significant performance increase with only four CI samples in $D_{few}$. This finding is consistent with the fact that solving $D_{test}$ requires four different compositional rule types, as shown in definition Table \ref{composition_tasks}. Similarly, $D_{test}$ from \textbf{Task2} and \textbf{Task3} require two and one samples illustrating required, but previously unseen compositional inference knowledge, respectively. We find most evaluated PLMs in Fig \ref{fig_few-shot}.b to drastically improve their performance with only two samples, and in Fig.\ref{fig_few-shot}.c with just a single sample. An exception is BERT, which requires more shots than the number of unseen inference cases. 

The above experiment suggests that the evaluated PLMs can greatly benefit from few-shot settings to enhance their systematicity capabilities. It is compelling that the number of samples in $D_{few}$ needed to reach substantial task performance corresponds to the number of inference knowledge types required to make correct inference predictions, i.e., making it possible to evaluate novel compositions. It will be interesting to study how to identify
potentially missing types of compositional inference knowledge for existing PLMs, and how to inject this knowledge in an efficient, data-free method.

\section{Conclusion}
We propose the first \textit{comprehensive} systematicity evaluation benchmark, SETI, applied to Natural Language Inference. Experiments on six widely used PLMs show that they can distinguish novel compositions with known primitives and composing knowledge with high accuracy, but limited when lacking such knowledge. Moreover, we show that models can quickly acquire missing inferential knowledge for systematicity by being presented with \textit{unique} samples representing each missing case of inferential knowledge, in a few-shot setup.

\section{Limitation}
SETI only considers veridical inference and natural inference (including both lexically-based inference and structure-based inference). However, our benchmark SETI can be flexibly extended to more varied reasoning patterns, such as negation, quantifiers, or others. In addition, we evaluate the systematicity capabilities of PLMs on semi-synthetic datasets, which are limited in language variance. Extending our benchmark on manually annotated compositional inference datasets might be a promising future work. \par
Recently, \citet{hupkes2020compositionality} dissect the notion of compositionality and define five theoretically grounded tests for generalization, in a task-agonistic manner. Our work is limited to evaluating the systematicity of PLMs in textual inference. While the systematicity test is one of the most important tests, the remaining ones (e.g., \textit{productivity} and \textit{localism}) are still worth to be explored in future works.

\section{Acknowledgments}
We are grateful to three anonymous reviewers for their valuable comments that have helped to improve this paper. This work has been supported through a scholarship provided by the Heidelberg Institute for Theoretical Studies gGmbH.

\bibliography{main.bbl}
\bibliographystyle{acl_natbib}

\clearpage
\appendix
\section{Veridical Inference}
\label{app:verbs}
In order to construct \textit{veridical inference}, we select 30 verbs, including 15 veridical verbs $f_{v}$ and 15 non-veridical verbs $f_{nv}$. Table \ref{tab_verbs} show instantiation of selected verbs.
\begin{table}[h]
\centering
\resizebox{0.98\columnwidth}{!}{
\begin{tabular}{p{2.1cm}p{7.5cm}} \toprule
Verb Types & Instantiations\\ \midrule
veridical verbs $f_{v}$ & realize, acknowledge, remember, note, find, notice, learn, see, reveal, discover, understand, know, admit, recognize, observe \\  \midrule
non-veridical verbs $f_{nv}$ & feel, claim, doubt, hope, predict, imply, suspect, wish, think, believe, hear, expect, estimate, assume, argue \\ \bottomrule
\end{tabular}}
\caption{Instantiation of veridical and non-veridical verbs used for constructing veridical inference.}
\label{tab_verbs}
\end{table}

\section{Data Stastics}
\label{app:statistics}
Since we use 30 verbs to construct premises $f_{v/nv}(s)$ for primitive veridical ($PI_{ver}$) and compositional ($CI$) inferences from the premises $(s)$ of natural inferences ($PI_{nat}$), the number of these two inference types is 30 times the amount of $PI_{nat}$, respectively. To avoid data biases in composition training, we guarantee the two major types from $D_{train}$ are balanced by downsampling the extensive inference type. For example, in Task1 trivial setting, we downsample $PI_{ver}$ to ensure the training data of $PI_{ver}$ and $PI_{nat}$ is balanced.

\section{Evaluated Pre-train Language Models}
\label{app:models}
We evaluate SETI across six well-known PLMs. Table \ref{tab_plms} shows the training objective and parameters of each model. Detailed information and training parameters of each model is:

\textbf{BERT} \cite{devlin-etal-2019-bert} is a bidirectional transformer pre-trained model, trained with masked language modeling and next sentence prediction objectives on a large corpus. We fine-tuned the base- uncased-large version, with the default setting.

\textbf{ALBERT} \citep{lan2020albert} build on BERT, and presents two parameter-reduction techniques to lower memory consumption and increase the training speed. We fine-tuned the ALBERT-large version, with the default setting.

\textbf{RoBERTa}  \cite{liu2019roberta} builds on BERT, but is trained without the next-sentence prediction objective and uses much larger data. We fine-tuned a RoBERTa-large version, with the default setting.

\textbf{GPT-2} \citep{radford2019language} is a decoder-only model, pre-trained on a large corpus of English data in a self-supervised fashion. We fine-tuned the GPT2-large version, with the default setting.

\textbf{BART} \citep{lewis-etal-2020-bart} is an encoder-decoder model. The pretraining task involves randomly shuffling the order of the original sentences and a novel in-filling scheme, where spans of text are replaced with a single mask token. We fine-tuned the BART-large version, with the default setting.

\textbf{T5} \cite{2020t5} is an encoder-decoder model which is pre-trained on a multi-task mixture of unsupervised and supervised tasks. Each task is in a text-to-text format. We fine-tuned the T5-large version, with the default setting.

\begin{table}[h]
\centering
\resizebox{0.98\columnwidth}{!}{
\begin{tabular}{llccc} \toprule
Model & Objective & Parameters & Layers & Type\\ \midrule
BERT & MLM+NSP &340M &24 & \multirow{3}{*}{Enc} \\ 
ALBERT & MLM+SOP &17M &24 \\ 
RoBERTa & MLM &355M &24\\ \midrule
GPT-2 & LM &774M &36 & Dec\\ \midrule
BART & DAE &406M &24 &\multirow{2}{*}{Enc-Dec}\\ 
T5 & DAE &770M &24\\ \bottomrule
\end{tabular}}
\caption{Overview of PLMs evaluated for systematicity in our work. For training objectives, \textit{MLM} is masked language modeling, \textit{NSP} is next sentence prediction objective, \textit{SOP} is sentence order prediction, \textit{LM} is language modele, and \textit{DAE} is the denoising autoencoder}.
\label{tab_plms}
\end{table}

\end{document}